\documentclass{sig-alternate-05-2015}
\usepackage{graphicx}
\usepackage{caption}
\usepackage{subcaption}
\usepackage{dblfloatfix}
\usepackage{fixltx2e}

\begin{document}






%

\title{Human-Algorithm Interaction Biases in the Big Data Cycle: A Markov Chain Iterated Learning Framework
\titlenote{This research was supported by National Science Foundation grant NSF-1549981.}
}
%
%
%
%
%

 \numberofauthors{2} 
 \author{
 \alignauthor
Olfa Nasraoui \\ 
       \affaddr{University of Louisville}\\
       \affaddr{Louisville, KY}\\
       \email{olfa.nasraoui@louisville.edu}
\alignauthor
Patrick Shafto \\ 
       \affaddr{Rutgers University -- Newark}\\
       \affaddr{Newark, NJ 07102}\\
       \email{patrick.shafto@rutgers.edu}
}

\maketitle
\begin{abstract}
Early supervised machine learning algorithms have relied on reliable expert labels to build predictive models. However, the gates of data generation have recently been opened to a wider base of users who started participating increasingly with casual labeling, rating, annotating, etc.  
The increased online presence and participation of humans has led not only to a democratization of unchecked inputs to algorithms, but also to a wide democratization of the ``consumption'' of machine learning algorithms' outputs by general users. Hence, these algorithms, many of which are becoming essential building blocks of recommender systems and other information filters, started interacting with users at unprecedented rates. The result is machine learning algorithms that consume more and more data that is unchecked, or at the very least, not fitting conventional assumptions made by various machine learning algorithms. 
These include biased samples, biased labels, diverging training and testing sets, and cyclical interaction between algorithms, humans, information consumed by humans, and data consumed by algorithms. 
Yet, the continuous interaction between humans and algorithms is rarely taken into account in machine learning algorithm design and analysis. In this paper, we present a preliminary theoretical model and analysis of the mutual interaction between humans and algorithms, based on an iterated learning framework that is inspired from the study of human language evolution. We also define the concepts of human and algorithm blind spots and outline machine learning approaches to mend iterated bias through two novel notions: antidotes and reactive learning.

\end{abstract}

%
%

\ccsdesc[500]{Computer systems organization~Embedded systems}
\ccsdesc[300]{Computer systems organization~Redundancy}
\ccsdesc{Computer systems organization~Robotics}
\ccsdesc[100]{Networks~Network reliability}


\section{Introduction}
\begin{figure}[!ht]
\begin{subfigure}[b]{0.5\textwidth}
        \includegraphics[scale=0.4]{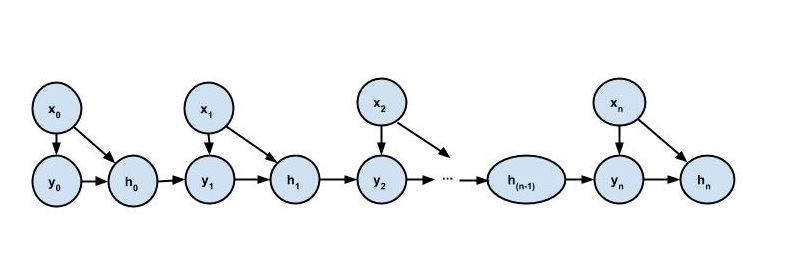}
        \caption{}
\end{subfigure}

\begin{subfigure}[b]{0.5\textwidth}
        \includegraphics[scale=0.4]{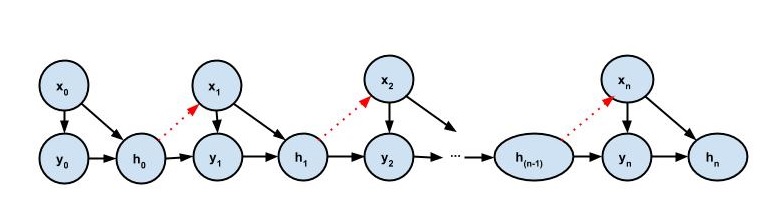}
                \caption{}
\end{subfigure}

\caption{\label{fig:Iterated-learning-figure} 
In iterated learning, information is passed through selected data, (a) 
where the inputs, $x$, are independent of the inferred hypothesis, and (b)
where the next inputs are selected based on the previous hypothesis. The latter case is more consistent with recommender systems and information filtering circumstances.
\normalsize
}
\end{figure}






Websites and online services offer large amounts of information,
products, and choices. This information is only useful to the extent that people can find what they are interested in. All existing approaches aid people by suppressing information that is determined to be unpreferred or not relevant. Thus, all of these methods, by gating access to information, have potentially profound implications for what information people can and cannot find, and thus what they see, purchase, and learn. 

There are two major adaptive paradigms to help sift through information: information retrieval and recommender systems. 
Information retrieval techniques \cite{sparck1970some,Robertson::1977,Jones::1978,Rijsbergen:1979,salton1983extended,Baeza-Yates::1999,Croft::2009} have given rise to the modern search engines which return relevant results, following a user's explicit query. For instance, in the probabilistic retrieval model \cite{Robertson::1977}, optimal retrieval is obtained when search results are ranked according to their relevance probabilities.
Recommender systems, on the other hand, generally do not await an explicit query to provide results \cite{Goldberg:1992:UCF:138859.138867,maes1994agents,Resnick::1997,Pazzani::1997,basu1998recommendation,Schafer::1999,Adomavicius::2005,Nasraoui::2006}.
Recommender systems can be divided based on which data they use and 
how they predict user ratings.
The first type is content-based filtering (CBF) algorithms  \cite{Pazzani::1997,Balabanovi::1997,pazzani2007content}. 
 It relies on item attributes or user demographics, but often not relations between users (i.e.\ social relations), as data. 
Collaborative Filtering (CF) \cite{Goldberg:1992:UCF:138859.138867,shardanand1995social,Konstan::1997,Sarwar::2001,Linden::2003,Nasraoui::2006}, on the other hand, does not require item attributes or user attributes. Rather it makes predictions about what a user would like based on what other similar users liked. 
Both adopt algorithms, e.g.\ K-nearest neighbors  \cite{Cover::1967,Dudani::1976} and non-negative matrix factorization (NMF)  \cite{Lee::1999,Koren::2009,Salakhutdinov::2008,Abdollahi::2014}, that have close analogs in the psychology literatures on concept learning, e.g.\ exemplar models \cite{medinschaffer78,nosofsky84,kruschke92} and probabilistic topic models \cite{griffiths_steyvers04,Griffiths07topicsin}.

Information filtering algorithms \cite{sheth1993evolving,maes1994agents,Hanani::2001} similarly provide users with a list of relevant results, but do so in response to a query. One classic example is the Rocchio filter  \cite{Rocchio::1971,buckley1995automatic,Pazzani::2007}, which modifies the user's initial
query after a first iteration of search to help filter less relevant results. The query is modified based on the set of initial search result documents which are labeled by the user as relevant and non-relevant, respectively. The new query (which is treated like a pseudo-document) is modified by adding and subtracting a weighted combination of relevant and non-relevant documents, respectively. This 
is quite similar to content-based recommendation, where information about the items is used to rank potentially relevant results. 

Common to both recommender systems and information filters is: \textbf{(1)} selection, of a subset of data about which people express their preference, by a process that is not random sampling, and \textbf{(2)} an iterative learning process in which people's responses to the selected subset are used to train the algorithm for subsequent iterations.
The data used to train and optimize performance of these systems are based on human actions. Thus, data that are observed and omitted are not randomly selected, but are the consequences of people's choices. 
Recommendation systems suggest items predicted to be of interest to a user (e.g. movies, books, news) based on their user profile  \cite{Resnick::1997,Pazzani::1997,Konstan::1997}. The prediction can be based on people's explicit (e.g. ratings) or implicit (e.g. their browsing or purchase history) data \cite{nasraoui2008web,khribi2008automatic,khribi2012automatic,zhuhadar2010hybrid}, or even query patterns \cite{zhang2006mining}. 
Research into human choice suggests that both explicit and implicit choices systematically vary based on context, especially the other options that are present when choosing
\cite{debreu60,tversky72,mcfadden76}.

In addition to the simple effects of the interaction between algorithms\textquoteright{}
recommendations and people\textquoteright{}s choices, people may reason
about the processes that underlie the algorithms. Research in cognitive
science has shown that people reason about evidence selected by other
people. In \cite{shaftogf12}, a computational framework was proposed for modeling
how people\textquoteright{}s inferences may change as a consequence
of reasoning about why data were selected. This framework has been formalized in learning from
helpful and knowledgeable teachers 
\cite{shaftogoodman08,bonawitzEtal11,buchbaumEtal11,shaftogg14}, deceptive informants \cite{warnerEtal11}, and epistemic trust \cite{shaftoEtal12,eavess2012,landrumes2015}. 
People's reasoning about the intentional nature of the algorithms may
exacerbate the effects of cyclic interaction between
the algorithms\textquoteright{} recommendations and people\textquoteright{}s
choices. 


We propose a framework for investigating the implications of interactions between human and algorithms, that draws on diverse literature to provide algorithmic, mathematical, computational, and behavioral tools for investigating human-algorithm interaction. 
Our approach draws on foundational algorithms for selecting and filtering of data from computer science, while also adapting mathematical methods from the study of cultural evolution \cite{griffithsk07,kirbyEtal07,beppu09CogSci} to formalize the implications of iterative interactions.

Key to our approach is the focus on the sources and consequences of bias in data collected ``in the wild''. The two primary sources of bias are from algorithms and from humans. Algorithms, such as recommender systems, necessarily filter information with the goal of presenting humans with typically the most preferred content. Then, based on the labels provided by people, learning algorithms are trained to optimize future recommendations. This framework differs from standard learning theory in that the training data are not randomly sampled, which calls into question any guarantees about learning from such data. The second source of bias is people. In addition to receiving filtered information optimized to their preferences, people are also not required to provide labels for any of the presented data. Moreover, people's choices are highly non-random, and may reflect not only their opinions about the presented content, but also inferences about why the content was presented. Finally, bias introduced into the data at any point may be magnified by retraining of models and associated implications for recommendations, yielding algorithms whose performance is at variance with theoretical expectations.  
We  argue that either of the individual sources of bias is in principle sufficient to yield instability, and that this suggests the need for new theories and methods for understanding performance of such systems in terms of human-algorithm interactions. 
We propose to characterize the conditions under which we would expect these to lead to systematic bias in the selection of information by algorithms, and identify conditions under which we can ``undo" the effects of these biases to obtain accurate estimates from biased data. We expect the results to contribute insights back to the fundamental psychology of human reasoning, choice and learning and the fundamental computer science of learning, recommendation and information filtering.





\section{Preliminary Analysis}
\subsection{Markov Chain Iterated Learning Analysis of Human-Algorithm Interaction}

In order to capture the iterative interaction between people and machine learning algorithms ``in the wild'', we could look into formal and empirical frameworks that have been developed in the behavioral sciences for analyzing the asymptotic effects of iterative interactions. For instance, we could consider the evolution of algorithms as a special case of cultural evolution of the sort observed in human language \cite{kirby01} and human knowledge more broadly \cite{tomasello00culturalOriginsBOOK,beppu09CogSci}. 
For this purpose, a formal framework is needed for analyzing the effects of local decisions on long-run behavior. One way to do this is using Markov chains to model iterative interactions with transmission between adjacent iterations as was done in \cite{griffithsk07}. While previous work concentrated on human behavior and learning, our focus here is on algorithm behavior and learning, namely in terms of the choice of data to present to people and the updated behavior in response to people's observed actions. 
More specifically, in the short term, we seek to identify the conditions under which we would expect algorithms' behavior to converge to more or less effective performance in the long run.
In the longer term, we seek to understand and devise mechanisms that can compensate for these biases to ensure good performance.

In our preliminary research, we start with simple supervised machine learning where the goal is to learn to predict a discrete class label. Research on simple classifiers has historically paved the road for most formal analysis in machine learning and data mining, and had a significant impact on both information retrieval \cite{sparck1970some} as well as recommender systems  \cite{basu1998recommendation}. It also provides a point of contact with the psychology of category learning (which we will exploit in future experiments). 


\subsubsection{Iterated learning with  \emph{filter-bias} dependency}

In the following, We extend the Markov Chain-based iterated learning framework to provide a framework to analyze the
evolution of the learned hypotheses \textbf{while taking into account interactions between the user and the algorithm}. 
One major extension from the original framework (see Fig \ref{fig:Iterated-learning-figure}(a)) is that we will explicitly take into account the dependency
between the current hypothesis learned by the algorithm (learner) and the next input supplied by the human. This is because, in each iteration, the model
or hypothesis that is learned by the algorithm can be considered to act as a filter
or gateway to the types of data that will later \textbf{be seen by}
the user.
This modification of the original graphical model will thus allow a dependence between the current hypotheses $h$ and the next inputs $\mathbf{x}$ (see Fig \ref{fig:Iterated-learning-figure}(b)).

The extent of the departure that we propose from a conventional machine learning framework toward a human - machine learning framework, can be measured by the contrast between the evolution of iterated learning \textit{without}
and \textit{with} the added dependency. Without the dependency, the algorithm at step $n+1$ sees input $\mathbf{x}_{n+1}$ which is generated
from a distribution $p(\mathbf{x})$ that is independent of all other
variables. 
Represent this independence with new notation $q(\mathbf{x})$ ($q$ instead of
$p$), 
where $q(\mathbf{x})$ represents an unbiased sample from the world, rather than a selection made by the algorithm. 
With the dependency, the algorithm at iteration $n+1$ sees
input $\mathbf{x}_{n+1}$ which is generated from a mixture between
the objective distribution $q(\mathbf{x})$ and another distribution that
captures the dependency upon the previous hypothesis $h_{n}$ which biases the future inputs seen by the user,

\[
p(\mathbf{x}_{n}|h_{n})=(1-\epsilon)p_{seen}(\mathbf{x}_{n}|h_{n})+\epsilon q(\mathbf{x}_{n}).
\]
In the case
of a rating based recommender or an optimal probabilistic information
filter \cite{Robertson::1977}, the probability 
of selecting the data is related to its rank. For a rating based recommender, the rank is based on the predicted rating, and for an optimal probabilistic information filter, the rank is based on the probability of relevance \cite{Robertson::1977}. In each case, the selection of $\mathbf{x}$ is based on whether it is likely to be highly rated or relevant (i.e. its corresponding $\mathbf{y}$ value), given $h$. 
Assume that  $y=1$ denotes the relevant class (0 otherwise). If $\mathbf{x}$ is chosen based on the probability of relevance,  $p(\mathbf{y}_n=1|\mathbf{x}_n, h_n)$, then
\begin{equation}
p_{seen}(\mathbf{x}_{n}|h_{n})=\dfrac{p(\mathbf{y}_{n}=1|\mathbf{x}_{n},h_{n})}{\sum_{\textbf{x}_{i}}p(\mathbf{y}_{i}=1|\mathbf{x}_{i},h_{n})}.
\label{eq:p-seen-of-x}
\end{equation}
the selection of inputs depends on the hypothesis, 
$p(\mathbf{x}|h_{n})\neq p(\mathbf{x})$, and therefore information is not unbiased, $p(\mathbf{x}|h_{n})\neq q(\mathbf{x})$.
The transition probabilities take into account (\ref{eq:p-seen-of-x}), and will be

\[
p(h_{n+1}|h_{n})=\sum_{\mathbf{x}\in X}\sum_{\mathbf{y}\in Y}p(h_{n+1}|\mathbf{x},\mathbf{y})p(\mathbf{y}|\mathbf{x},h_{n})p_{seen}(\mathbf{x}|h_{n}).
\]

This can be used to derive the asymptotic behavior of the Markov chain
with transition matrix $T(h_{n+1},h_{n})=p(h_{n+1}|h_{n})$, 

\begin{equation}
p(h_{n+1})=\epsilon p(h_{n+1})+(1-\epsilon)T_{bias} \label{X}
\end{equation}

\noindent
where, 

\begin{equation}
T_{bias} =
\left[
\sum_{\mathbf{x}\in X}
\sum_{\mathbf{y}\in Y}
p(h_{n+1}|\mathbf{x},\mathbf{y}) 
\sum_{h_{n}\in H} 
p(\mathbf{y}|\mathbf{x},h_{n}) \\
p_{seen}(\mathbf{x}|h_{n})
\right] 
p(h_{n}). 
\label{Y}
\end{equation}

\noindent
Thus, iterated learning with a filter bias converges to a mixture of the prior and the bias induced by filtering. 
To illustrate the effects of the filter bias, we can analyze a simple and most extreme case where the filtering algorithm shows only the most relevant data in the next iteration (e.g.\ top-1 recommender). Hence  

\begin{equation}
x^{top}=\arg\max_{x}\left(p(y|\mathbf{x},h)\right), 
\label{eq:top1-filter-bias}
\end{equation}

\[
p_{seen}(\mathbf{x}_{n}|h_{n})=\begin{cases}
1 & for\: x=x^{top}=\arg\max_{x}\left(p(y=1|\mathbf{x},h)\right)\\
0 & otherwise,
\end{cases}
\]

\[
T_{bias}=
\left[\sum_{\mathbf{x}\in X}\sum_{\mathbf{y}\in Y}\left(\dfrac{p(\mathbf{y}|\mathbf{x},h_{n+1})p(h_{n+1})}{p(\mathbf{y}|\mathbf{x})}\right)\sum_{h_{n}\in H}p(\mathbf{y}|\mathbf{x}_n^{top},h_{n})\right]p(h_{n}).
\]

\noindent
The fact that $\mathbf{x}_n^{top}$ maximizes $p(y|\mathbf{x},h)$ suggests limitations to the ability to learn from such data. Specifically, the selection of relevant data allows the possibility of learning that an input that is predicted to be relevant is not, but does not allow the possibility of learning that an input that is predicted to be irrelevant is actually relevant. In this sense, \textbf{selection of evidence based on relevance is related to the confirmation bias in cognitive science}, where learners have been observed to (arguably maladaptively) select data which they believe to be true (i.e.\ they fail to attempt to falsify their hypotheses) \cite{klaymanh1987}. \textbf{Put differently, recommendation algorithms may induce a 
{\em blind spot} where data that are potentially important for understanding relevance are never 
seen}. 

------------------------------------------------------------

\subsubsection{Iterated learning with \emph{active-bias} dependency}

Active learning is a classical method in machine learning \cite{Cohn::1994,cohn1996active,Settles::2010,Castelli::1995,bruneretal56}, used to reduce the number of labeled samples required for learning and thus accelerate learning versus training sample addition. Starting with a seed set of labeled instances, typically, future training inputs are selected and added to the training data based on hypotheses to {\em improve} learning. 
Consider the case of one classical approach in active learning \cite{Settles::2010}, where
data are selected to be presented to the user for labeling based on
the uncertainty portended by its prediction using the current algorithm's
hypothesis, 

\begin{equation}
p_{active}(\mathbf{x}|h)\propto1-p(\mathbf{\hat{y}}|\mathbf{x},h).
\label{eq:p-active-of-x-prop}
\end{equation}
where 
$\mathbf{\hat{y}}=\arg\max_{y}\left(p(y|\mathbf{x},h)\right)$. %
That is, $\mathbf{x}$ values are selected to be least certain about $\mathbf{\hat{y}}$, the predicted y. 

Considering a simplified algorithm where only the very best data are selected, we can investigate the limiting behavior of an algorithm with active learning bias. Assuming a mixture of random sampling and active learning, we obtain:
\[
x^{act}
= \arg\max_{x} \left(1-p(\mathbf{\mathbf{\hat{y}}}|\mathbf{x},h)\right), 
\]

\begin{equation}
p(h_{n+1})=\epsilon p(h_{n+1})+(1-\epsilon)T_{active}, \label{Z}
\end{equation}

\begin{equation}
T_{active}=\left[\sum_{\mathbf{x}\in X}\sum_{\mathbf{y}\in Y}p(h_{n+1}|\mathbf{x},\mathbf{y})\sum_{h_{n}\in H}p(\mathbf{y}|\mathbf{x}_n^{act},h_{n})\right]p(h_{n}). \label{W}
\end{equation}

The limiting behavior depends on the active learning bias, $T_{active}$. 
When the active-bias algorithm is certain about the rating or relevance of an item, it will never select it. In contrast, the filtering algorithm is almost certain to pick items that it knows are relevant. Thus, the goals of active learning and filtering are in opposition; filtering tends to select items for which the prediction is certain (to be highly rated or relevant), whereas active learning selects items for which prediction is uncertain. 

This is, of course, consistent with the different goals of recommendation and active learning. 
The analysis illustrates how the long-run implications of these different biases may be analyzed: By deriving the transition matrices implied by iterated application of data selection biases, we can see that both active learning and filtering have different goals, but focus on an ever more extreme (and therefore not representative) subset of data. Similar methods can be applied to more nuanced and interesting biases to shed light on the consequences of iterative interactions on the data. 


\subsection{Iterated learning with human action bias}\label{ssec:humanBias}

The above analysis 
assumes that people's response is always observed. 
In the following, we extend our analysis to the more realistic case where users have a choice of whether to act or not on a given input. 

Assume that people have some target hypothesis, $h^*$, which represents optimal performance for the algorithm. 
Data are composed of an input provided by the algorithm, $x$, an output, $y$, and an action, $a$. 
The indicator variable $a$ takes a value of $1$ when people have provided a response, and a value of $0$ when people have not. 
When the value of $y$ is not observed, it is notated as $y=$ \textsc{null}. 
These form triples $\mathbf{d}=(\mathbf{x}, \mathbf{y}, \mathbf{a})=\{(x_1,y_1,a_1), ..., (x_n,y_n, a_n)\}$. 
The basic inference problem, from one iteration to the next, is then,
\[
p(h_t|\mathbf{d}) \propto p(\mathbf{d}|h_{t-1}, h^*)p(h_{t-1}),
\]

\begin{equation}
p(h_t|\mathbf{d}) \propto  p(\mathbf{y}|\mathbf{x},\mathbf{a},h^*)
p(\mathbf{a}|\mathbf{y^*},\mathbf{x},h^*)
p(\mathbf{x}|h_{t-1})p(h_{t-1}),
\end{equation}
where $\mathbf{y}^*$ represents the output that would be observed, if an action were taken. 
The main change is in people's choice of whether to respond, $p(\mathbf{a}|\mathbf{y^*},\mathbf{x},h^*)$. 
A missing at random assumption implies that $p(\mathbf{a}|\mathbf{y^*},\mathbf{x},h^*)$ does not depend on $\mathbf{x}$, $\mathbf{y}^*$, or $h^*$, thus $p(\mathbf{a}|\mathbf{y},\mathbf{x},h^*) = p(\mathbf{a})$. If variables are missing due to a person's choice, the probability of a missing value almost certainly depends on $\mathbf{x}$, $\mathbf{y}^*$, and/or $h^*$.
We can formalize this choice using Luce choice \cite{luce59}, a special case of softmax \cite{suttonbarto}, 
\footnote{Both softmax and Luce choice have known issues for modeling human choice \cite{debreu60,reiskampbm06}.}
\begin{equation}
p(\mathbf{a}=1|\mathbf{y^*},\mathbf{x},h^*) = \frac{U(\mathbf{a}=1|\mathbf{y^*},\mathbf{x},h^*)}
{U(\mathbf{a}=0|\mathbf{y^*},\mathbf{x},h^*) +
U(\mathbf{a}=1|\mathbf{y^*},\mathbf{x},h^*)},
\label{eq:humanChoice}
\end{equation}
where the choice of whether to act depends on the relative utility of acting as opposed to not acting. 
For example, if it is especially effortful to act, then people will be biased against acting. 
Alternatively, the utility of acting may depend on the value of $\mathbf{y}^*$. For example, it may be that there is greater perceived utility in acting when the value of $\mathbf{y}^*$ is very low, as in the case of an angry customer or disappointed user.

In principle, one might think that this is related to the problem of dealing with missing data that is common in statistics \cite{Rubin::1976}. Indeed, in our analyses, we showed one special case that reduces to the missing at random typically assumed in statistical applications \cite{Rubin::1976}. However, the framework proposed here is in fact more general; it proposes a theory of {\em why} data are missing, and formalizes the problem as one of understanding human behavior \cite{shaftogf12,shaftob2015,durkincbs2015}.


\subsection{A Blind Spot to Learning and to Human Exploration}\label{ssec:bubble}

In the following, we present 
formal definitions that allow 
quantifying the extent and influence of interaction bias on humans and algorithms. We start with defining the concept of blind spot relative to a human interacting with an algorithm.


$\delta^{H}_{b}$ --- \textbf{Human Blind Spot:} This is the set of data items from a universe of existing items or data records, $D$, available to a relevance filter algorithm, for which the probability
of being seen 
by the human interacting with the algorithm that has so far learned hypothesis $h$,
is less than or equal to $\delta^{H}_{b}$, 
\begin{equation}
\mathbf{D_{\delta^{H}_{b}}^{B}}=\{\mathbf{x} \in \mathbf{D} \mid p_{seen}(\mathbf{x}|h)\leq\delta^{H}_{b}\}. \label{eq:blindSpot}
\end{equation}

Analogous concepts and metrics can be defined from the perspective of the algorithm instead of the human, by replacing the probability that an item is seen or discoverable by the human, by the probability that it is observed, particularly along with a relevance label, by the algorithm.

$\delta^{A}_{b}$ --- \textbf{Algorithm Blind Spot:} The set of data available to a relevance filter algorithm,  for which the probability
of being seen along with a label provided by a human, 
by the algorithm that has so far learned hypothesis $h$, while interacting with that human,
is less than or equal to $\delta^{A}_{b}$, 
\begin{equation}
\mathbf{D_{\delta^{A}_{b}}^{B}}=\{\mathbf{x} \in \mathbf{D} \mid p(\mathbf{a}=1|\mathbf{y^*},\mathbf{x},h^*)  \leq\delta^{A}_{b}\}. \label{eq:blindSpot}
\end{equation}

Different learning biases may lead to different levels of blind spot prevalence relative to humans or to algorithms, as can be quantified below.

\textbf{Blind Spot Prevalence:} 
is defined as the \emph{proportion} of data in the blind spot, relative to the human:
\begin{equation}
\rho^{H}_{b}=\left|\mathbf{D_{\delta^{H}_{b}}^{B}}\right|/\left|\mathbf{D}\right|, \label{eq:Proportion}
\end{equation}
or relative to the algorithm:
\begin{equation}
\rho^{A}_{b}=\left|\mathbf{D_{\delta^{A}_{b}}^{B}}\right|/\left|\mathbf{D}\right|. \label{eq:Proportion}
\end{equation}

Metrics associated with the blind spot generated by a particular learning algorithm and bias generated,  
can help track the effect of human-algorithm interaction
with more iterations under various conditions of data
dependency, data selection, type of relevance filter algorithms,
type of human choice of action, different initialization bias, etc.

\subsection{Undoing the effects of bias}\label{ssec:undo}


We propose two possible approaches for undoing bias: 
\textbf{Antidotes} and \textbf{Reactive learning}. An \textit{antidote} mends the bias after it has  occurred, i.e. post-learning, for instance by adding tolerance to relevance boundaries. 
On the other hand, \textit{reactive learning} is an extension of active learning, where learning is altered by \textit{reacting} to the human or algorithm bias-induced selection strategies by intervening within the iterated learning process.

\subsubsection{Antidotes}\label{ssec:antidote}

There are several ways we can pursue an antidote; for example, by unbiasing the final predicted outputs or by using ensemble-based active learning such as \cite{Melville::2004}. Unbiasing can be performed post-learning by a reverse-Rocchio approach which works in the opposite way to traditional Rocchio personalization \cite{Rocchio::1971}. Here, we use selected data to change the set of relevant and non relevant instances. The tuning of the constant multiplier weights in Rocchio can affect the final status of filtering and severity of blind spots ($\delta_{b}$).
Also, personalized ensembles may be learned, such that they operate with different blind spots or blind spot levels to allow fast recovery and adaptation in filtering strength. 
One method is to block the data, treating nearby items as having a common parameter, and modeling the data generation process as a sequential process, using the inferred models for each block. This would allow us to simulate data from the different blocks of the process. 

\subsubsection{Reactive Learning}\label{ssec:Reactive}

Reactive learning can be achieved by incorporating an un-biasing strategy in each iteration of learning.
Reactive bias can span the entire range of selection mechanisms ranging from active learning to filter-bias and inverse-filter bias, including inverse sampling (note however, that existing methods like \cite{vardi1985} do not assume iterated learning). 
Reactive learning can leverage using models of human behavior that can allow us to explore, by simulation, how combinations of the previously explored algorithms may or may not lead to blind spots and bias. Some alternative strategies, such as reverse filter-bias, can also be simulated using a model of human behavior developed for filtering. 
Un-biasing an iterated human-machine learning mechanism can be approached by equalizing the selection bias in active learning. One way this has been explored in semi-supervised learning, where the availability of labels can be biased, is by inverse sampling which multiplies the sampling probability by the reciprocal of the probability of labels given data \cite{vardi1985,Cacoullos:1966}. However, care must be taken not to adversely affect the intended benefits of personalized information filters and recommender systems. To help unbias filters, we can explore adapting (into reactive learning) Active Collaborative Filtering strategies, such as \cite{boutilier2002active} 
and Uncertainty Sampling with Diversity Maximization (USDM) \cite{yang2014multi}.




\subsection{Comparison with existing work and limitations}

Researchers have been aware that data, fed to filters such as recommender systems, is biased by the mechanism by which users rate items \cite{herlocker2004evaluating}. Viewing the learning resulting from the interaction between an algorithm and a human as a dynamic process instead of a process that works on one static batch of data is reminiscent of dynamic machine learning, but the latter is a completely different concept which is more related to a paradigm for learning models under an evolving or dynamic data input. The way we consider dynamics here, however, is different. In fact, our analysis is not situated within a learning paradigm that is intended to learn a final predictive model while coping with dynamic data. Rather, we are in the context of analyzing the emergence of biases and related phenomena such as human and algorithm blind spots, when the learning algorithm receives data from humans while these humans receive predictions from the algorithm in an iterated cycle.  
We are also interested in studying the impacts of these biases and related phenomena on human and algorithm learning during the process of online machine learning.

Within the context of online machine learning from human activity data, dynamic usage patterns were studied in \cite{nasraoui2008web} and dynamic recommender systems were studied within a stream data mining framework in \cite{nasraoui2007performance}. In \cite{saka2010recommender}, a Swarm Intelligence-based recommender system, inspired by the collaborative behavior of bird flocks and called FlockRecom, generated recommendations by iteratively adjusting the position and speed of dynamic flocks of agents in a virtual space. Even with taking previous work, including all the aforementioned work into account, no prior work has studied human-algorithm interaction via iterated learning mechanisms. Even common benchmark data sets used in recommender systems are rife with biases. In fact, most past recommenders worked with or collected  sparse rating data and none has considered iterated learning-induced biases, but rather only corrected for simple models of user, item, or time based biases \cite{Koren::2009}. 
The latter type of bias, time based recommendation bias \cite{Koren::2009}, is not related to algorithm biases, but rather to general \textit{temporal} trends of movie preferences that are decoupled from the machine learning algorithm itself. Experiments with different methods to select the items to be rated before any recommendations \cite{rashid2002getting} showed different bias in initial ratings resulting from different methods. These item selection methods, which at the time, were limited to selecting items based on their distribution in the accumulated rating data so far, included popularity, random, entropy or personalized. Finally, the experiments simulating the various item selection methods started with a full rating matrix from other users, which is likely biased and could not capture biases 
resulting from 
iterated interaction between human and machine learning. In addition to these biases, previous conventional recommender systems do not record the full set of recommendations made in each iteration. This has consequently created a gap in the current public data sets that are available for benchmarking new algorithms. To fill the experimental benchmarking gap that limits using public data to study iterated human-algorithm biases, it is necessary to record each recommendation, user response, and the order, alongside past user history.

Our work may seem related to various aspects of active learning \cite{Cohn::1994,cohn1996active,Settles::2010,Castelli::1995,bruneretal56}, semi-supervised learning \cite{mitchell2006learning,nigam2006semi,chapelle2006semi,Zhu::2007,Zhu::2009,abdullin2012semi}, label sampling bias in semi-supervised learning \cite{rosset2004method}, and missing data theory \cite{Rubin::1976,Scheffer::2002}. 
One type of active learning, known as Proactive Learning \cite{donmez2008proactive} attaches a cost to every sample before asking for labels and assumes that different oracles have different reliability. However, none of these approaches perform a formal iterated learning analysis, nor an interdisciplinary study based on both human and machine learning, let alone behavioral experiments which we plan to undertake in our ongoing work. 
In our research so far, we have made formal preliminary analyses and simulations that already illustrate the differences between some of these related concepts and our questions. Specifically, none of the previous research has studied the iterated human-algorithm interaction as we have proposed, {\em through iterated learning between algorithms and humans} for the special case of filtering models, whether they be content-based or collaborative. For instance some work tried to study the effect of missing data in CF \cite{ma2007effective,marlin2009collaborative}; however no prior studies exist using an iterated model or measuring the impact of a machine learning algorithm itself on future data. Instead, the data is assumed to be available as one frozen batch, then analysis is performed based on that image.

Our notion of a blind spot is related to the concept of a filter bubble. Filter bubbles have been a subject of attention lately \cite{Pariser::2011,Brossard::2013,Liao::2013}, including work attempting to reverse or ameliorate such effects \cite{Melo::2013,Munson::2010}. Our formal definition of blind spot paves the way to formalize the definition of filter bubbles. Interestingly, filter bubbles can be thought of as the opposite of blind spots, where the information is nearly certain to be seen within some small number of recommendations. This is consistent with the idea that a filter bubble prevents mixing of ideas, and suggests that our Markov chain-based analyses may provide tools to analyze the implications of different recommendations on filter bubbles. 
Our approach differs in that we focus on developing a framework for understanding the emergence of biases and filter bubbles, dynamic changes, and long run behavior of algorithms. Beyond mending filter bubbles, we bring forward a combination of mathematical, algorithmic, computational and behavioral perspectives that promises to yield insight into when and why blind spots and filter bubbles emerge, and how different algorithms are likely to behave over iterations. 

Research in psychology has investigated category learning \cite{bruneretal56,shepardetal61,posnerKeele68,medinschaffer78,AshbyAlfonsoReeseWaldron98}, choice behavior \cite{thurstone27,luce59,yellott77,tversky72,simonsont1992}, active learning \cite{bruneretal56,AshbyQuellerBerretty99,AshbyAlfonsoReeseWaldron98,ashbymb2002,markant2014select}, and social learning \cite{shaftogoodman08,bonawitzEtal11,buchbaumEtal11,shaftogg14}. Perhaps the nearest neighbor is the literature on the implications of other people for learning. This literature has shown that people reason about {\em why} other people select data, and use people's selection of data to update their beliefs about the data selection process \cite{shaftoEtal12,eavess2012,landrumes2015}. Our approach builds from insights drawn from each of these works, and thus embodies an extension of traditional work in psychology to the theoretically interesting and practically important domain of human-algorithm interaction.

\section{Conclusion and Future Outlook}

There is a long tradition in machine learning of algorithms whose performance is guaranteed in the context of unbiased data. 
Similarly, there is a long tradition in the psychology of human learning of treating learning as inference from unbiased data. Increasingly, people and algorithms are engaged in interactive processes wherein neither the humans nor the algorithms receive unbiased data. What are the long-run consequences of these iterative interactions on algorithms' performance? On human knowledge?
The unique contributions of this preliminary research arise from bringing together mathematical, algorithmic, behavioral, and computational perspectives from computer science and psychology to understand how algorithm performance and human behavior depend on one an other, and how those dependencies affect long run performance. 
Our ongoing work will pave the road for a framework on which the study of human-algorithm interaction may progress. 


%
\bibliographystyle{abbrv}
\bibliography{references}  
%

\end{document}